\documentclass[lettersize,journal]{IEEEtran}
\usepackage{amsmath,amsfonts}
\usepackage{algorithmic}
\usepackage{array}
\usepackage[caption=false,font=normalsize,labelfont=sf,textfont=sf]{subfig}
\usepackage{textcomp}
\usepackage{stfloats}
\usepackage{url}
\usepackage{verbatim}
\usepackage{color,xcolor}
\usepackage{colortbl} 
\usepackage{graphicx}
\usepackage{caption}
\usepackage[mathscr]{euscript}
\usepackage{amssymb}
\usepackage{multirow}
\usepackage[normalem]{ulem}

\useunder{\uline}{\ul}{}
\def\BibTeX{{\rm B\kern-.05em{\sc i\kern-.025em b}\kern-.08em
    T\kern-.1667em\lower.7ex\hbox{E}\kern-.125emX}}
\usepackage{balance}
\begin{document}
\title{From Canteen Food to Daily Meals: Generalizing Food Recognition to More Practical Scenarios}
\author{Guoshan~Liu,
        Yang~Jiao,
        Jingjing~Chen, \textit{Member, IEEE},
        Bin~Zhu and
        Yu-Gang~Jiang \textit{Fellow, IEEE}
\thanks{This work was supported by Shanghai Science and Technology Program Project (No. 21JC1400600), and partially supported by the Singapore Ministry of Education (MOE) Academic Research Fund (AcRF) Tier 1 grant. Jingjing Chen is the corresponding author.\\
\indent Guoshan Liu, Yang Jiao, Jingjing Chen and Yu-Gang Jiang are with Fudan Vision and Learning Laboratory (FVL), Fudan University, Shanghai, China (e-mail: \{22210240029, yangjiao21\}@m.fudan.edu.cn, \{chenjingjing, ygj\}@fudan.edu.cn). Bin Zhu is with Singapore Management University, Singapore(e-mail: binzhu@smu.edu.sg).}
}

\markboth{IEEE TRANSACTIONS ON MULTIMEDIA}%
{How to Use the IEEEtran \LaTeX \ Templates}

\maketitle

\begin{abstract}
The precise recognition of food categories plays a pivotal role for intelligent health management, attracting significant research attention in recent years. Prominent benchmarks, such as Food-101 and VIREO Food-172, provide abundant food image resources that catalyze the prosperity of research in this field. Nevertheless, these datasets are well-curated from canteen scenarios and thus deviate from food appearances in daily life. This discrepancy poses great challenges in effectively transferring classifiers trained on these canteen datasets to broader daily-life scenarios encountered by humans. Toward this end, we present two new benchmarks, namely DailyFood-172 and DailyFood-16, specifically designed to curate food images from everyday meals. These two datasets are used to evaluate the transferability of approaches from the well-curated food image domain to the everyday-life food image domain. In addition, we also propose a simple yet effective baseline method named Multi-Cluster Reference Learning (MCRL) to tackle the aforementioned domain gap. MCRL is motivated by the observation that food images in daily-life scenarios exhibit greater intra-class appearance variance compared with those in well-curated benchmarks. Notably, MCRL can be seamlessly coupled with existing approaches, yielding non-trivial performance enhancements. We hope our new benchmarks can inspire the community to explore the transferability of food recognition models trained on well-curated datasets toward practical real-life applications.
\end{abstract}


\section{Introduction}

\begin{figure}[!t]
\centering
\includegraphics[width=3.2in]{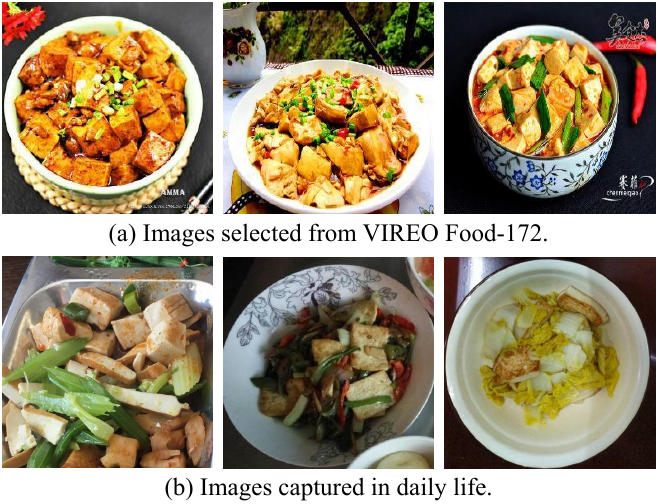}
\caption{Variations in visual appearance of ``Braised Tofu" from VIREO Food-172 and daily meals. The first row shows three examples of dishes in VIREO Food-172, followed by examples from daily meals in the second row.}
\label{fig1}
\vspace{-0.2in}
\end{figure}

\IEEEPARstart{F}{OOD} computing has developed remarkably with the rise of deep learning \cite{oxford, Cedric, Image}, which facilitates people to monitor and regulate their nutritional intake and caloric consumption. Several pioneering benchmarks \cite{oxford, Deep-based, food101, Region-Wise} offer abundant food images resources, where Food-101 \cite{food101}, Food2K \cite{oxford} are mainly consist of western food, while Vireo Food-172 \cite{Deep-based} and Vireo Food-251 \cite{Region-Wise} are composed of Chinese food.
To ensure aesthetics and discernibility, these datasets collect food images from dishes cooked by professionals in canteen scenarios. This data collection behavior introduces inherent bias into the curated datasets. Since culinary experts master extensive recipe knowledge and superb cooking skills, they tend to follow similar cooking procedures and use similar ingredients when preparing the same dishes. Consequently, this bias leads to a reduced variance among samples within the same type of dishes. Nevertheless, dishes in daily meals generally exhibit higher variance due to the non-standard cooking procedures and the randomness of shooting angles. 
As shown in Fig.\ref{fig1}, we compare the images of the same dish ``Braised Tofu" from Vireo Food-172 and daily meals in the first and second rows, respectively. First, the comparison of examples within each row reveals that the same type of dishes from Vireo Food-172 exhibit high appearance consistency, while those from daily life vary dramatically. Besides, dishes from Vireo Food-172 are more exquisite than their daily-life counterparts.
Therefore, classifiers pre-trained on these canteen food datasets struggle to effectively generalize to daily meals, thereby limiting their broader applicability.

To the best of our knowledge, there is no available dataset that encompasses food images captured in daily meals at present, which hinders the development of practical food recognition.
To mitigate this gap, we curate two new benchmarks, namely DailyFood-172 and DailyFood-16, which include 172 and 16 categories of daily-life food images respectively. The DailyFood-172 is collected from the ``Go Cooking"    \footnote{
https://www.xiachufang.com/category/} website, where culinary enthusiasts share their everyday dishes. On the other hand, DailyFood-16 gathers images from ``Eggplant Health"\footnote{https://www.qiezilife.com/}, a platform for weight loss purposes. Due to the cumbersome and resource-intensive nature of collecting and annotating a vast number of daily-life food images, merely utilizing the proposed datasets as extensional training resources still limits the scalability of models. Therefore, in this paper, we employ our datasets as the benchmark for evaluating the transferability of existing approaches. In this scenario, deep models have access to labeled food images from existing benchmarks such as Food-101 and Vireo Food-172, while our DailyFood-172 and DailyFood-16 datasets only provide unlabeled images during training. 

With the above restrictions, our benchmarks can be formulated as an Unsupervised Domain Adaptation (UDA) problem, where deep models are trained with both label-rich source dataset and label-scarce target dataset. By mitigating the distribution discrepancy between source and target domains, pioneering domain-level UDA approaches \cite{MMD, DAN, CORAL, DSN, Similarity-based} have demonstrated improved generalization abilities. 
Another line of works \cite{CCUDA, DSAN, DAAN, CAN} explores learning the pseudo labels for target data to address the category-level shift. In this paradigm, the pseudo-label prediction has been extensively studied \cite{pseudo, DADA} as it determines the domain shift aiming to learn within each category. Approaches leveraging fine-grained alignment for category-level domain shift learning often demonstrate superior generalization compared to those focused solely on domain-level shifts

Despite their effectiveness, existing approaches fall short of achieving satisfactory generalization results due to the inherent variability present in the target domain of daily-food images. Compared to existing food datasets, our DailyFood series exhibit a notable characteristic of significant intra-class variance, leading to the challenge of ``category ambiguity". As illustrated in the last image of the second row in \ref{fig1}, the assigned category label ``Braised Tofu" can be ambiguous to distinguish from another class ``Shredded cabbage" due to some shared ingredients. Previous methods \cite{DSAN, DAAN, DADA} have primarily focused on learning single-category cross-domain alignment, which is considered a sub-optimal solution for our datasets.
In light of this, we propose a simple yet effective baseline method called Multi-Cluster Reference Learning (MCRL). Our MCRL relaxes the restriction that image features from the target domain should be pulled closer to one specific source cluster while being distant from other clusters in the source domain. Specifically, MCRL first predicts pseudo-labels for each sample in the target domain dataset and groups them into distinct semantic clusters following previous methods \cite{DADA, MEDA, JDA}. Subsequently, MCRL encourages samples within these clusters to simultaneously learn distribution shifts towards multiple source cluster features dynamically. This dynamic learning enables the model to adaptively weigh the relevance of each target sample to multiple source clusters during training. Through this learning approach, MCRL not only addresses the aforementioned ``category ambiguity" problem but also mitigates the adverse effects of imperfect pseudo-label predictions. Furthermore, it is noteworthy that MCRL can be seamlessly integrated with existing UDA methods, resulting in significant performance improvements.

We summarize our contributions as following three folds:
\begin{enumerate}
\item{We introduce two new high-quality daily-life food recognition benchmarks DailyFood-172 and DailyFood-16 to unlock the potential of transferring food recognition models trained on well-curated canteen food datasets to the daily life scenario. 
}
\item{We propose a simple yet effective baseline method named Multi-Cluster Reference Learning (MCRL) for our DailyFood-172 and DailyFood-16 datasets. Within our MCRL, multiple source clusters are dynamically referred to comprehensively learn the domain gap.} 
\item{We extensively evaluate the transferability of existing approaches on our proposed DailyFood-172 and DailyFood-16. Besides, we couple our proposed MCRL with several state-of-the-art UDA approaches, and comprehensive performance enhancements prove the effectiveness of our proposed method.}
\end{enumerate}


\section{RELATED WORK}

{\bf{Dataset for Food Recognition. }} Over the years, the research community has developed several high-quality food datasets. For instance, the ETH Food-101 dataset compiled by \cite{food101} encompasses a collection of 101,000 images distributed across 101 Western food categories. VIREO Food-172 \cite{Deep-based} consists of 110,241 images from 172 Chinese food categories. FoodX-251 \cite{FoodX-251} has also been frequently utilized for classification tasks. In addition, there are other food-relevant large recipe datasets, such as Food2K \cite{oxford} and Recipe1M \cite{Cooking}. Food2K \cite{oxford} and Recipe1M \cite{Cooking} are representative of large-scale food-related datasets but with two important differences. First, the primary purpose of Recipe1M is to facilitate cross-modal embedding and retrieval between recipes and images, whereas Food2K is tailored towards propelling scalable food visual feature learning. Secondly, Recipe1M predominantly encompasses over 1 million structured cooking recipes, each linked to a series of food images. Conversely, Food2K is an aggregation of more than 1 million images, falling under 2,000 food categories. Based on these datasets, many works related to food have emerged as a topic of significant interest. These efforts include, but are not limited to, food quantity estimation based on depth images \cite{Automatic}, image segmentation for volume estimation \cite{Recognition}, multi-food recognition \cite{Multiple-food}, and food image recognition \cite{FOODCLASSIFICATION, Database}. Nonetheless, the majority of existing datasets for food recognition are either composed of restaurant food photos or collected from cooking enthusiast websites. Most of these images depict food that has undergone relatively professional cooking, plating, and photography, which significantly differs from the appearance of food images captured in daily life primarily for documenting meals. To evaluate the performance of food recognition models in real-life scenarios, we constructed two food datasets, namely DailyFood-172 and DailyFood-16, which closely resemble everyday usage scenarios.

{\bf{Unsupervised Domain Adaptation. }} While food recognition has recently captured numerous research attention, cross-domain food recognition has been less studied. 
To the best of our knowledge, there is only one work that investigates the problem of cross-domain food computing in the literature \cite{food-transfer}. 
This work proposes a cross-domain cross-modal food transfer network that employs adversarial learning to reduce the gap between source and target domains for recipe retrieval tasks. As our work focuses on unsupervised cross-domain adaptation for food recognition, therefore, we review recent work on unsupervised cross-domain adaptation. 

Current unsupervised domain adaption (UDA) methods basically make alignment between source and target domain either at domain-level \cite{MMD, DANN, DSN, Return} or in category level. Domain-level UDA minimizes the distribution divergence between the source and target domain by pulling them into the same distribution at different scale levels. The majority of the approaches rely on discrepancy minimization and adversarial training. The former minimizes the discrepancy between domains using a statistical distance function. With the introduction of MMD (Maximum Mean Discrepancy) \cite{Ghifary}, \cite{DAN, JAN} used MMD to learn more transferable features for domain adaptation. In the context of adversarial training, supervision is provided by a learned domain discriminator in a GAN \cite{GAN} framework to encourage domain-invariant features \cite{Backpropagation, Dlow, Cycada, DANN}. In the past few years, several works focus on the fine-grained category-level label distribution alignment \cite{pseudo, DTN, CAN, MADA}. Compared to coarse-grained alignment at the domain level, such kind of methods attempt to align each category distribution between the source and target domain, by using the pseudo label \cite{Pseudo-label} as a guide, to improve the classification accuracy. Generally, previous methods for category-level alignment involve aligning the conditional distribution \cite{CAN, Spherical, Fisher}. To increase the robustness and accuracy of pseudo labels, \cite{Rectifying} utilizes the prediction variance of two classifiers as an estimation of pseudo label uncertainty, prioritizing samples with low prediction variances for increased reliability. \cite{pseudo} uses a spanning tree to delete low-confidence samples to train the model. 
However, the aforementioned methods do not adequately adapt classification models trained on the fine-dining food dataset to everyday food images. Therefore, we propose a baseline method MCRL to address this issue.
\begin{figure*}[!t]
\centering
\includegraphics[width=\textwidth]{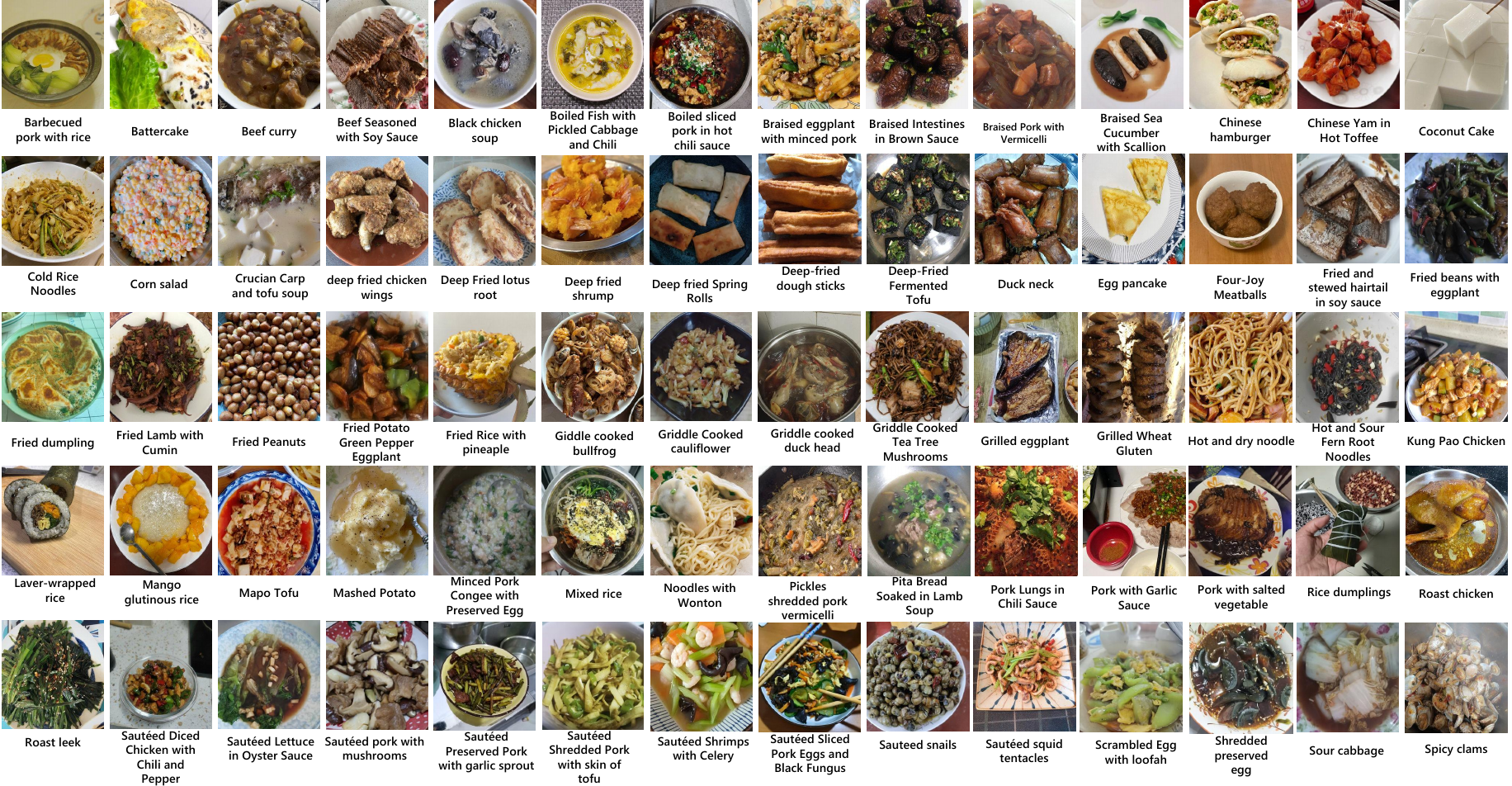}
\caption{Examples of food categories in DailyFood-172.}
\label{DailyFood-172_pickup}
\vspace{-0.2in}
\end{figure*}

\vspace{-0.1in}
\section{DATASET CONSTRUCTION}
We construct two curated datasets DailyFood-172 and DailyFood-16\footnote{https://github.com/SLKAlgs/DailyFood-172\label{web}}, which are made publicly available, both of which consist of food images captured from daily life dishes.
 
\subsection{DailyFood-172}
Since the UDA problem requires the homogeneity of categories of the source and the target domain, our newly constructed dataset, DailyFood-172, has categories that perfectly align with the VIREO Food-172 dataset, which consists of the 172 most common Chinese dishes. The images in the DailyFood-172 dataset are crawled from the ``Go Cooking" \footnote{https://www.xiachufang.com/category/} website. On this website, each dish category has several corresponding recipes, each with a unique ingredient list and cooking procedure. Users have uploaded numerous images for each recipe. Considering that variations in ingredient composition and cooking procedures can lead to differences in food appearance, we carefully select images from every recipe for each dish category to ensure the diversity of our curated dataset. Subsequently, we resize each image to a resolution of 280×280 pixels. Following this selection process, our DailyFood-172 dataset now contains 42,312 images, with minimum 47 and maximum 329 samples per category. Fig.\ref{DailyFood-172_pickup} showcases some examples of food categories included in DailyFood-172. For a detailed distribution of the sample count per category, please refer to the DailyFood-172 GitHub repository\textsuperscript{\ref {web}}. On average, there are 246 samples per category.

\subsection{DailyFood-16}
\begin{figure}[!t]
\centering
\resizebox{0.4\textwidth}{!}{\includegraphics{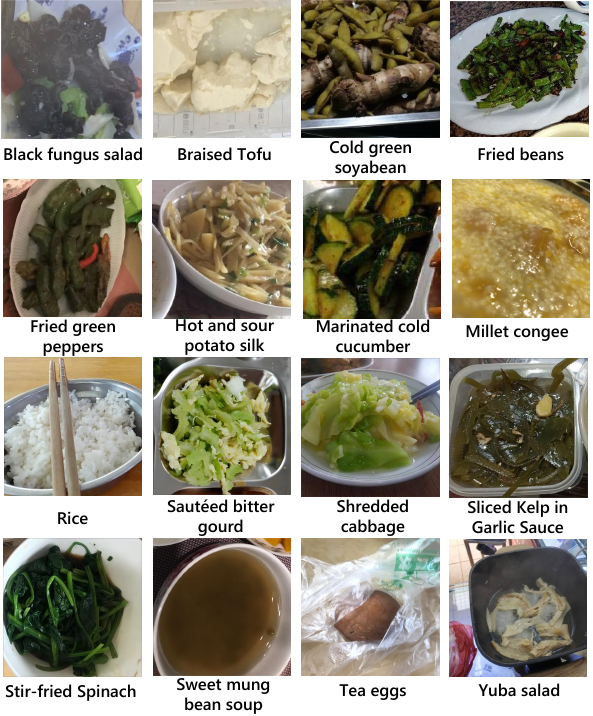}}
\caption{Examples of food categories in DailyFood-16.}
\label{DailyFood-16_pickup}
\vspace{-0.2in}
\end{figure}

\begin{figure}[!t]
  \centering
  \resizebox{3.5in}{!}{\includegraphics{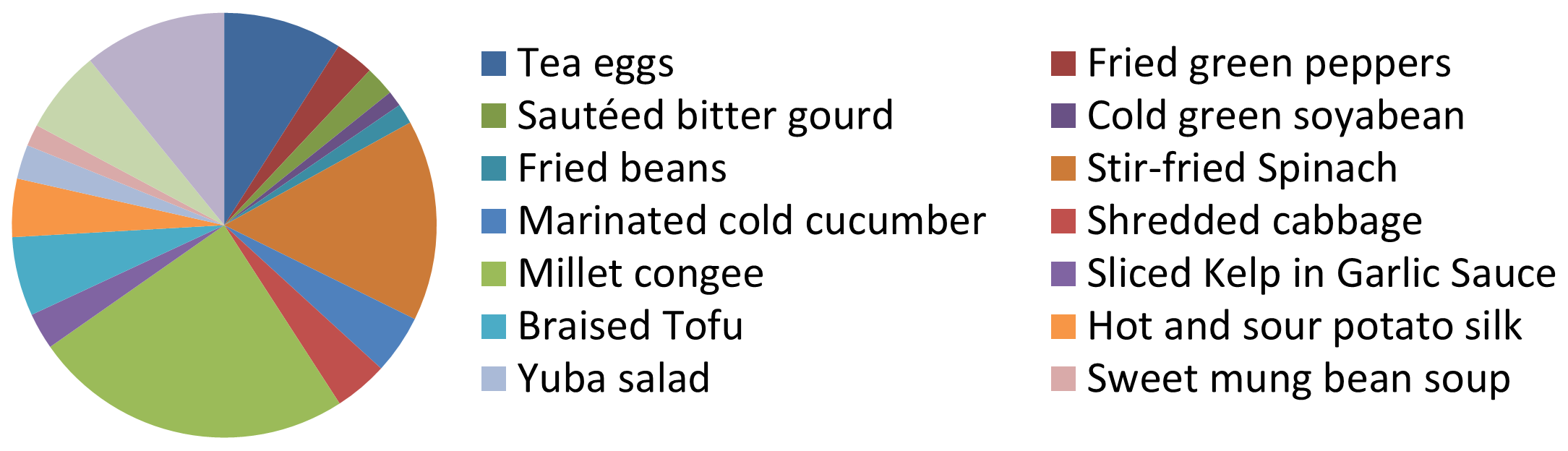}}
  \caption{Pie distribution chart of categories in DailyFood-16.}
  \label{category}
\vspace{-0.2in}
\end{figure}

Different from DailyFood-172, DailyFood-16 is collected from the diet recording platform ``eggplant health" which is specialized for users who want to lose weight. On this platform, users are required to upload pictures of the food they consume each meal. These images are used for analysis by nutritionists to provide advice. Therefore, the photos on this platform serve as a means of fulfilling the task rather than being intended for sharing purposes. As a result, these pictures often do not prioritize aesthetic appeal. We construct DailyFood-16 following the subsequent steps. First, since the uploaded images only have ingredient descriptions rather than category labels, we manually label a subset of images with the predefined labels in VIREO Food-172 \cite{Deep-based} according to ingredient overlappings. For instance, if the user's descriptions include ``spinach", we will annotate it as ``spinach" related dishes, e.g., ``stir-fried spinach".
However, such an annotating manner will lead to the one-to-many problem, i.e., a certain ingredient can correspond to multiple dishes. Toward this end, we require multiple annotators to manually assign the most likely category label from multiple possible choices for each image according to their consensus and select the label obtaining the most votes. Lastly, considering that the uploaded images often include cluttered backgrounds, which will disturb the model's attention, we crop the food-relevant part to guarantee the quality of our dataset. With the above steps, our DailyFood-16 includes 1,695 images in total with minimum 20 and maximum 413 samples per category and covers 16 food categories in VIREO Food-172. In DailyFood-16, each food category approximately contains 106 samples on average. Due to the irregularity of image cropping, the sizes of images in the DailyFood-16 dataset are different. Fig.\ref{DailyFood-16_pickup} shows some examples of food categories in DailyFood-16 and Fig.\ref{category} presents the distribution of sample numbers in each category. 

It is worth mentioning that although the ingredients can provide more fine-grained cues for food recognition and such resources are also available in both VIREO Food-172 and our DailyFood-16, we do not formulate our DailyFood-16 as an ingredient-level UDA benchmark because the ingredients from DailyFood-16 exhibit high variety and randomness, exceeding the ranges of those in the VIREO Food-172, which deviates the traditional UDA problem. Therefore, we only use the category labels as ground truths in our DailyFood-16 benchmark. 

\section{METHOD}
In this section, we will first preliminarily introduce the problem definition of Unsupervised Domain Adaptation (UDA) and the basic formulation of the majority of the existing solutions. Then, we will elaborate on the motivation as well as detailed designs of our proposed Multi-Cluster Reference Learning (MCRL) method. 



\subsection{Preliminary}
In Unsupervised Domain Adaptation (UDA), a source domain 
$\mathcal{D}_{\mathit{s}}=\left\{x^{s}_{i}, y^{s}_{i}\right\}^{n_{s}}_{i=1}$ of $n_{s}$ labeled examples and a target domain $\mathcal{D}_{\mathit{t}}=\left\{x^{t}_{j}\right\}^{n_{t}}_{j=1}$ of $n_{t}$ unlabeled examples are generally defined.
$y^{s}$ are labels of samples in the source domain. The purpose of UDA is to transfer the model trained on source domain $\mathcal{D}_{\mathit{s}}$ to the target domain $\mathcal{D}_{\mathit{t}}$ without catastrophic performance loss by mitigating the domain shift between source and target domains.
As summarized in Section \uppercase\expandafter{\romannumeral2}, existing leading methods, i.e., category-based UDA solutions, focus on obtaining a good alignment of the source and target domains in a class-wise manner. Specifically, a feature extractor $g_{\theta}$ and a classifier $f_{\theta}$ can be trained on the source domain using labeled data as:

\begin{equation}
\begin{aligned}
    & f_{i} = g_{\theta}(x_i) \\
    & L_{y} = -\frac {1}{n_{s}}\sum _{x_{i}\in{\mathcal{D}_{\mathit{s}}}}\sum _{c=1}^{C}P_{x_{i}\rightarrow c} \log f_{\theta}(f_{i}),
\end{aligned}
\end{equation}
where C is the number of classes, $P_{x_{i}\rightarrow c}$ is the probability of $x_{i}$ belonging to class c.
Afterward, $f_{\theta}$ is then used to generate pseudo labels $\hat{y}^t$ for target domain samples $\mathcal{D}_{\mathit{t}}=[x_1^t,...,x_{n_t}^t]$ as:
\begin{equation}
\hat{y}^t = \mathop{argmax}\limits_{c}(\mathrm{Softmax}(f_{\theta}(f_{i}))),
\end{equation}
we define the features of source samples $X_{\mathit{s}}=\left\{x^{s}_{i}\right\}^{n_{s}}_{i=1}$ as $F_{s}=\{f^{s}_{i}\}^{n_{s}}_{i=1}$,
Subsequently, the labels $Y_{\mathit{s}}=\left\{y^{s}_{i}\right\}^{n_{s}}_{i=1}$ are employed to partition the feature $f_i^S$, yielding class-specific clusters:
\begin{equation}
    C_i^{(c)}=\mathrm{MERGE}(F_{s}^{(c)}),
\end{equation}
and for each sample $x^{s}_{i}$ with the corresponding ground truth $y^{s}_{i}$, the index operator $[y^{s}_{i}]$ can be used to extract the feature of the corresponding category. Then, the features of the same category $c$ are merged to obtain the cluster. The cluster for each category is represented as $F_s^{(c)}$:
\begin{equation}
    F_s^{(c)}=\{(f_i^s,y_i^s),y_i^s=c\},
\end{equation}
here, $f_i^s$ indicates the feature of the $i$th sample in its corresponding category $c$, while $F_s^{(c)}$ denotes the features corresponding to category $c$. 
Then, the shift between the target domain and its corresponding category within the source domain is minimized by learning the domain-invariant features between the target samples and the source domain clusters. This process aims to achieve maximal alignment between the source and target domains within the feature space. Numerous methods exist for the computation of distances between two distributions, including but not limited to Cosine Distance, Kullback-Leibler Divergence, and Maximum Mean Discrepancy(MMD) \cite{kernel}. 
Formally, MMD defines the following difference measure based on category: 




\begin{equation}
d_{\mathcal{H}}(p,q)\triangleq \mathbf{E}_c\left \| {\mathbf{E}_{p^{(c)}}}\phi(x^s) - {\mathbf{E}_{q^{(c)}}}\phi(x^t)\right \|_{\mathcal{H}}^2,
\end{equation}
where $x^s$ and $x^t$ are the instances in $\mathcal{D}_s$ and $\mathcal{D}_t$, and $p^{(c)}$ and $q^{(c)}$ are the distributions of $\mathcal{D}_s^{(c)}$ and $\mathcal{D}_t^{(c)}$ respectively. 
By minimizing Equation (4) in deep networks, the distributions of the same category are drawn close.
\vspace{-0.1in}

\begin{figure}
  \centering
  \begin{minipage}{0.24\textwidth}
    \centering
    \includegraphics[width=\linewidth]{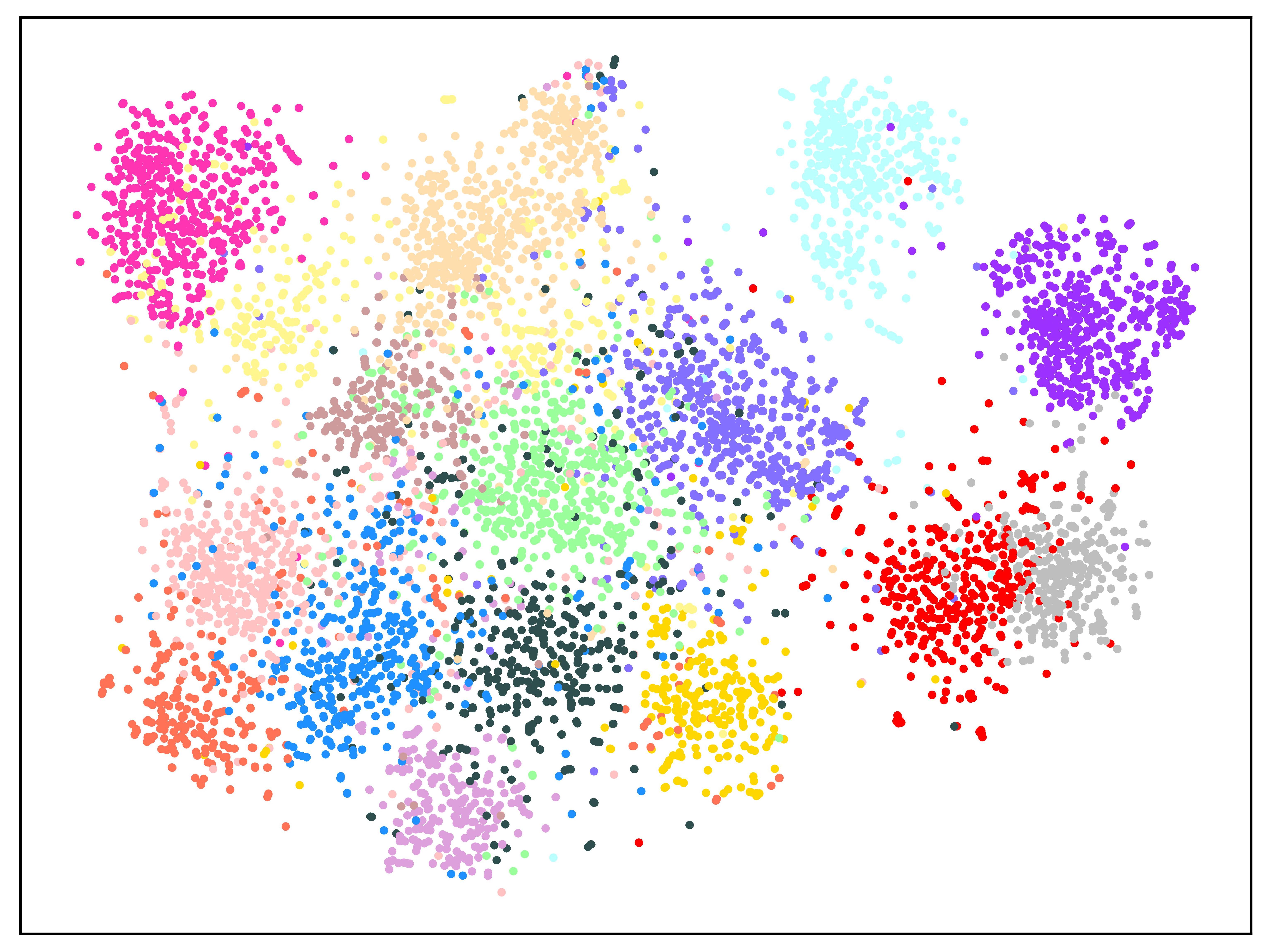}
    \caption*{(a) Feature distributions of images in VIREO Food-172.}
    \label{tsne1}
  \end{minipage}
  \hfill
  \begin{minipage}{0.24\textwidth}
    \centering
    \includegraphics[width=\linewidth]{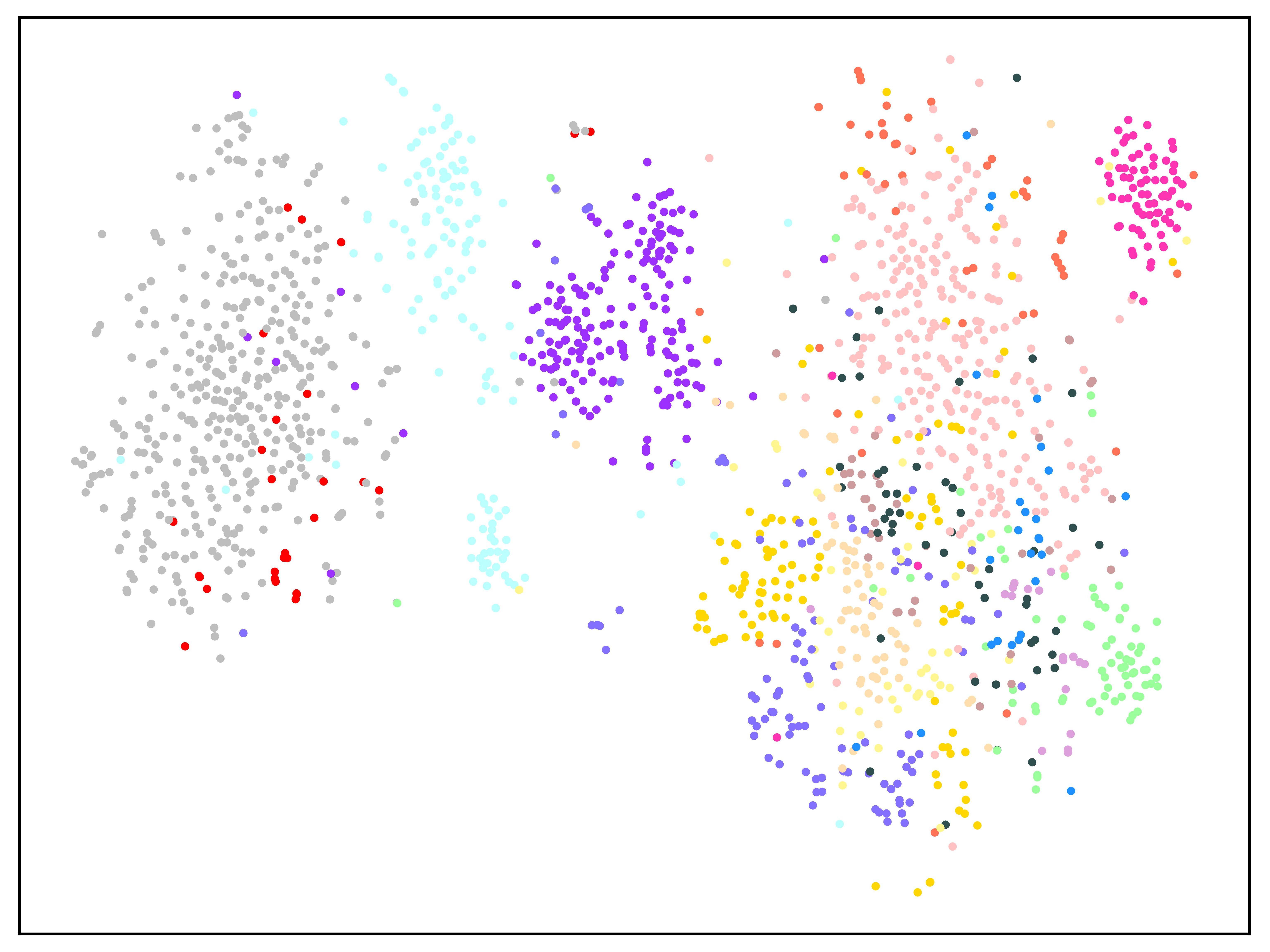}
    \caption*{(b) Feature distributions of images in DailyFood-16.}
    \label{tsne3}
  \end{minipage}
  \caption{(Best viewed in color) The t-SNE visualization of features in different datasets. (a) and (b) respectively represent the feature distributions extracted from the VIREO Food-172 \cite{Deep-based} and DailyFood-16 dataset, using the ResNet-50 \cite{resnet} as backbone. (Colors identical are utilized for the same categories.)}
  \label{tsne}
\vspace{-0.2in}
\end{figure}

\subsection{Multi-Cluster Reference Learning}
\begin{figure*}[!t]
\centering
\resizebox{0.87\textwidth}{!}{\includegraphics{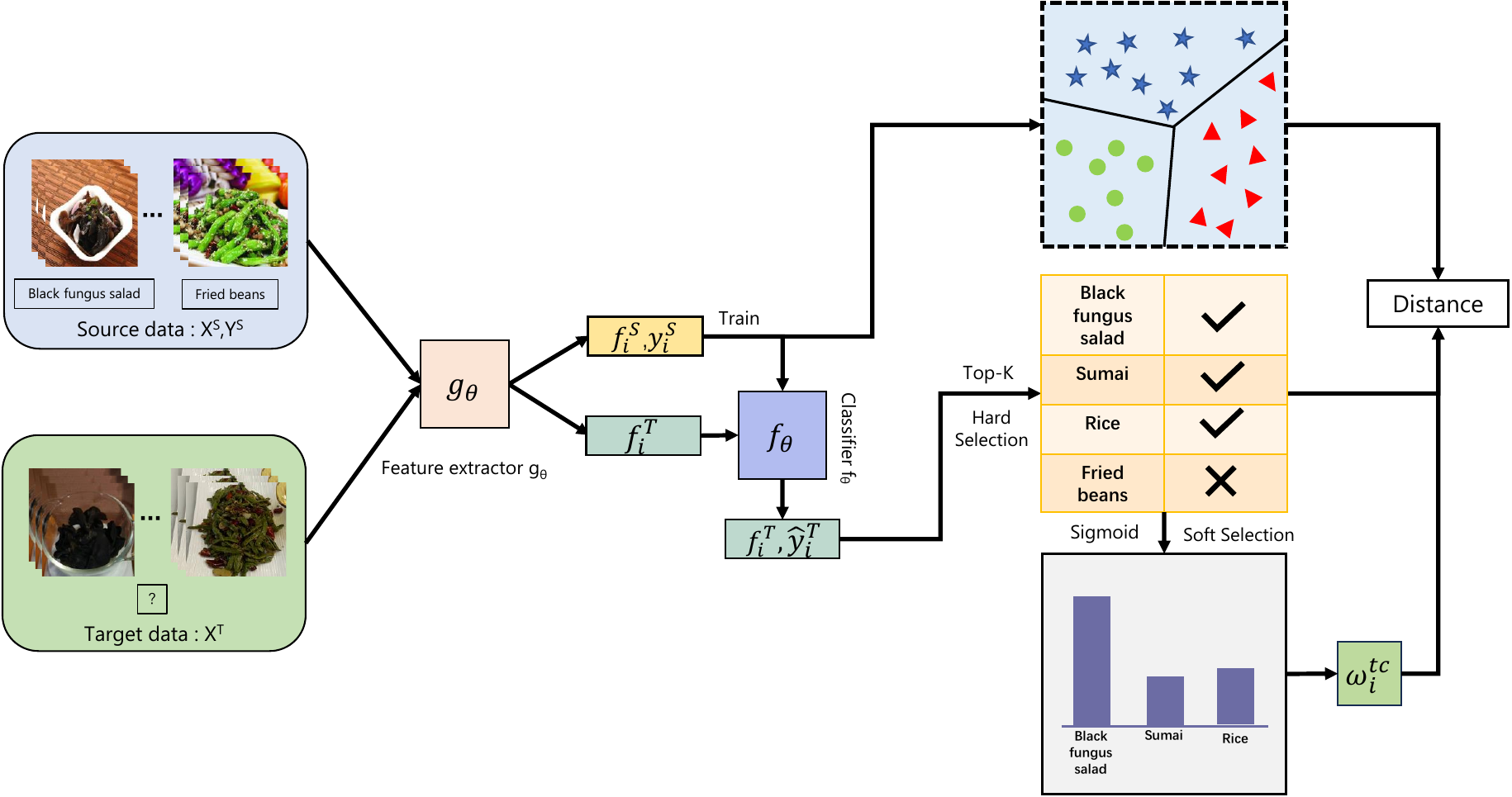}}
\caption{\textbf{The architecture of the proposed Multi-Cluster Reference Learning.} It consists of a feature extractor $g_\theta$, a classifier $f_\theta$, as well as two multi-cluster reference learning objectives. Within the two learning objectives, the distribution gap between target samples and multiple categories of source clusters is narrowed for pursuing better generalization ability.} 
\label{framework}
\end{figure*}

As illustrated in Fig \ref{tsne}, taking the t-SNE visualizations of VIREO Food-172 (Fig. \ref{tsne}(a)) with the same categories in DailyFood-16 (Fig. \ref{tsne}(b)) dataset as examples, compared to the standardized features in VIREO Food-172 \cite{Deep-based}, the data in the DailyFood-16 dataset shows a state of large intra-class feature variations and unclear inter-class feature differences, which leads to the problem of ``category ambiguity'' mentioned in Section \uppercase\expandafter{\romannumeral1}. To solve this issue, we propose the Multi-Cluster Reference Learning (MCRL) framework, as shown in Fig. \ref{framework}. It simultaneously learns the domain gap between a certain target sample and multiple source clusters. 

In the proposed MCRL, the features obtained from the source data through $g_\theta$ are used for clustering, where samples of the same category are gathered around the same clusters. According to Equation (2), the category corresponding to the highest probability following the Softmax operation of the classifier $f_\theta$ has traditionally been used as the pseudo label for the sample. However, due to the ambiguity and diversity of ingredients among food images, we choose $K$ categories with the highest probabilities as the pseudo labels for each target sample. Consequently, each target sample can correspond to multiple category clusters for reference. Finally, the distribution shifts between the target domain samples and several source domain clusters are minimized during training.
We employed two strategies when deciding the target clusters, i.e., MCRL with hard and soft selection, which will be elaborated on in the following parts.

\subsubsection{MCRL with Hard Selection}
Compared to Equation (2), we now aim to obtain the $K$ categories with the highest probability after the Softmax function. First, we calculate the probability of each target sample $P=\{p_1,...,p_C\}$ as:
\begin{equation}
P = \mathrm{Softmax}(f_{\theta}(f_i^T)),
\end{equation}
where $f_i^T$ denotes the features obtained from the target domain samples through the feature extractor $g_\theta$. 
Then we select the $K$ highest probability of $P$ and then obtain their corresponding pseudo labels. To align the distribution of the target samples with cluster features $F_s^{(K)}$ of these $K$ categories in the source domain, we compute the distribution discrepancy between the target domain samples and the source domain clusters as learning targets:



\begin{equation}
d_{\mathcal{H}}(p,q)\triangleq\frac{1}{K}\sum_{k=1}^K\left \| F_s^{(k)}-\frac{1}{|\mathcal{D}_t|}\sum_{x_j^t\in{\mathcal{D}_t}}\phi(x_j^t)\right \|_{\mathcal{H}}^2,
\end{equation}
where $F_s^{(K)}$ are the feature clusters of the $\mathrm{Top-K}$ categories corresponding to each target sample in the source domain and $\phi(x_j^t)$ is the distribution of target sample $x_i^t$. In this way, for each target domain sample, we can simultaneously reduce its domain shifts from the multiple source clusters that correspond to its categories. For each $\mathrm{Top-K}$ category, we utilize identical weights to reduce the discrepancy between domains to learn distribution shifts.
\subsubsection{MCRL with Soft Selection}
Considering the varying resemblance between individual target samples and source categories, it is reasonable to encourage the model to adaptively control the magnitude of domain shift learning with different source category clusters. Therefore, we also devise a dynamic scoring mechanism that enables the model to learn weights for controlling the domain shift learning process.
To avoid the adverse effect of introducing too much noise when setting the learning target, we posit that a single image most likely correlates with the features of a maximum of $K$ categories. Consequently, we initially procure the probability $\hat{Y}^t$ of the $\mathrm{Top-K}$ categories by leveraging equations (6). 
Inspired by the fact that the sigmoid function is more suitable for multi-label classification tasks \cite{qin2019rethinking}, we generate $K$ controlling weights with sigmoid functions as:

\begin{equation}
w_j^{tc}=\mathrm{Sigmoid}(f_{\theta}(f_i^T)), w_j^c={w_j^1,...,w_j^K},
\end{equation}
where $w_j^{tc}$ denotes the individual probabilities of the target sample $x_j^t$ among $c$ categories, and $c\leq K$. 
Accordingly, equation (7) is modified as follows:


\begin{equation}
d_{\mathcal{H}}(p,q)\triangleq\frac{1}{K}\sum_{k=1}^K\left \| F_s^{(k)}- \frac{1}{|\mathcal{D}_t|}\sum_{x_j^t\in{\mathcal{D}_t}}w_j^{tc}\phi(x_j^t)\right \|_{\mathcal{H}}^2,
\end{equation}
where $w_i^{sc}$ denotes the weights of each sample within the $K$ selected categories corresponding to the target samples in the source domain, and $w_j^{tc}$ denotes the weight of $x_j^t$ belonging to class $c$. 
Hard selection, assigning equal weights to each sample, is a specific case of soft selection, which accounts for the uncertainty of target samples.

In summary, by incorporating the controlling scores of the $K$ selected categories as weights into the original category-based UDA method, our MCRL method can dynamically learn and enable the model to adaptively balance the relevance of each target sample to multiple source clusters during the training process. With this learning approach, MCRL not only solves the previously mentioned ``category ambiguity" problem but also alleviates the negative impacts brought by imprecise pseudo-label prediction. 

\section{EXPERIMENT}
\subsection{Implementation Details}

Unlike typical UDA datasets like Office-Home \cite{officehome} and VisDA \cite{VISDA} with multiple domains, we use the VIREO Food-172 training set \cite{Deep-based} as the source domain and evaluate on two target datasets DailyFood-172 and DailyFood-16. 

\begin{table}[]
\centering
\caption{Performance comparison. ``Source-only" denotes that the classifiers are pre-trained on VIREO Food-172 without using any UDA approaches.}
\label{comparision}
\resizebox{0.5\textwidth}{!}{
\begin{tabular}{c|c|c|c}
\hline
Backbone                   & Methods         & DailyFood-172(\%)               & DailyFood-16(\%)                \\ \hline
\multirow{6}{*}{ResNet50}  & Source-only     & 66.62                           & 74.34                           \\
                           & DSAN\cite{DSAN}            & 66.96                           & 82.24                           \\
                           & DAAN\cite{DAAN}            & 68.26                           & 79.06                           \\
                           & DANN\cite{DANN}            & 67.41                           & 79.47                           \\
                           & DeepCoral\cite{CORAL}       & 66.36                           & 75.04                           \\
                           & CoVi\cite{covi}            & 73.82                           & 85.25                           \\ \hline
\multirow{5}{*}{ResNet101} & Source-only     & 67.48                           & 78.41                           \\
                           & DSAN\cite{DSAN}            & 68.04                           & 81.53                           \\
                           & DAAN\cite{DAAN}            & 68.48                           & 80.00                           \\
                           & DANN\cite{DANN}            & 67.49                           & 84.37                           \\
                           & DeepCoral\cite{CORAL}       & 67.14                           & 78.11                           \\ \hline
\multirow{5}{*}{ResNet152} & Source-only     & 67.95                           & 78.88                           \\
                           & DSAN\cite{DSAN}            & 68.25                           & 80.94                           \\
                           & DAAN\cite{DAAN}            & 69.22                           & 81.24                           \\
                           & DANN\cite{DANN}            & 67.67                           & 79.82                           \\
                           & DeepCoral\cite{CORAL}       & 67.89                           & 74.16                           \\ \hline
\multirow{2}{*}{ViT}       
                           & CDTrans\cite{CDTrans} & 81.93                           & 90.86                           \\
                           & CDTrans+MCRL (Ours) & \textbf{82.10} & \textbf{93.63}
                           \\ \hline
\end{tabular}
}
\vspace{-0.2in}
\end{table}


We implement all deep models using the PyTorch framework and utilize ImageNet \cite{imagenet} pre-trained ResNet-50, ResNet-101, and ResNet-152 models \cite{resnet} as backbone networks. Note that the labels for the target domain are solely used for evaluation purposes. To validate the effectiveness of our method, we integrate it into three algorithms: Deep Subdomain Adaptation Network (DSAN) \cite{DSAN}, Dynamic Adversarial Adaptation Network (DAAN) \cite{DAAN}, and Cross-Domain Transformer (CDTrans) \cite{CDTrans}. For DSAN and DAAN, we employ mini-batch stochastic gradient descent (SGD) with a momentum of 0.9 and a learning rate of 0.01. Throughout the experiments, we fix the epoch at 20 and the batch size at 32. As for CDTrans, we utilize the Adam optimizer \cite{adam} with a momentum of 0.9 and a learning rate of 0.0003. Similarly, we keep the epoch at 40 and the batch size at 32. CDTrans specifically employ the Vision Transformer (ViT) \cite{vit} as its backbone. For initializing our model, we used backbones trained from Data-efficient Image Transformers (DeiT) \cite{deit}, namely DeiT-Small (DeiT-S) and DeiT-Base (DeiT-B). To ensure a fair comparison, we set the parameters of the extended methods to be the same as those of the original methods. The evaluation metric used is the classification accuracy in the target domain.

\subsection{Performance of Multi-Cluster Reference Learning}

To examine the effectiveness of our proposed datasets DailyFood-172 and DailyFood-16 when used as target domains, for image classification UDA, we combine our method with the state-of-the-art method CDtrans. To further illustrate the effectiveness of MCRL, we compared CDTrans+MCRL which used DeiT-B as the backbone with several conventional UDA methods, which are DSAN \cite{DSAN}, DAAN \cite{DAAN}, DANN \cite{DANN}, DeepCoral \cite{CORAL}, CoVi \cite{covi}, and CDTrans \cite{CDTrans}, respectively. To ensure a fair comparison, the results for the comparative methods were sourced from their public codes or directly cited from the original papers. The TABLE \ref{comparision} lists the comparison results on the DailyFood-172 and DailyFood-16 datasets. It is not hard to observe that CDTrans+MCRL achieves the best performance in all tasks. This indicates that, as it currently stands, our method exhibits the best performance in terms of domain adaptation for food image recognition.

MCRL can be easily combined with many UDA algorithms. Taking into consideration that accuracy alone cannot provide a comprehensive evaluation of the model's performance, the macro-F1 score is employed to assess the overall balance between classification precision and recall of the model. TABLE \ref{f1} lists the comparative performances of the Top-1, Top-3 accuracies and Macro-F1 scores of DSAN \cite{DSAN}, DAAN \cite{DAAN}, CDTrans \cite{CDTrans} (DeiT-S), and CDTrans (DeiT-B) before and after the extended of our method, with the DailyFood-172 and DailyFood-16 datasets serving as the target domains respectively, while VIREO Food-172 \cite{Deep-based} is always the source domain. From the table, we can see that, after incorporating MCRL, the performance of DSAN, DAAN, and CDTrans has been significantly improved. To be specific, in terms of Top-1 accuracy, the majority of the baselines exhibited varying degrees of improvement after employing the MCRL. Specifically, the improvement rates for CDTrans using DeiT-S backbone, CDTrans with DeiT-B backbone, and DSAN with ResNet50 as the backbone on the DailyFood-16 dataset were 3.11\%, 2.77\%, and 2.18\% respectively. Eventually, in the vast majority of cases, these three metrics witnessed varying degrees of improvement after the application of MCRL. This demonstrates that MCRL can be effectively integrated into existing UDA methods, not only achieving a good cross-domain distribution alignment on food-related cross-domain datasets but also perfectly resolving the issue of ``category ambiguity" mentioned before.
\vspace{-0.1in}

\begin{table*}[]
\centering
\caption{Performance comparison on target datasets DailyFood-172 and DailyFood-16.}
\label{f1}
\begin{tabular}{c|c|ccc|ccc}
\hline
\multirow{2}{*}{Backbone} &
  \multirow{2}{*}{Methods} &
  \multicolumn{3}{c|}{DailyFood-172} &
  \multicolumn{3}{c}{DailyFood-16} \\ \cline{3-8} 
 &
   &
  \multicolumn{1}{c|}{Top-1(\%)} &
  \multicolumn{1}{c|}{Top-3(\%)} &
  Macro-F1(\%) &
  \multicolumn{1}{c|}{Top-1(\%)} &
  \multicolumn{1}{c|}{Top-3(\%)} &
  Macro-F1(\%) \\ \hline
\multirow{4}{*}{ResNet50} &
  DSAN\cite{DSAN} &
  \multicolumn{1}{c|}{66.96} &
  \multicolumn{1}{c|}{83.43} &
  65.35 &
  \multicolumn{1}{c|}{82.24} &
  \multicolumn{1}{c|}{95.75} &
  71.90 \\
 &
  \cellcolor{lightgray}\textbf{+MCRL(Ours)} &
  \multicolumn{1}{c|}{\cellcolor{lightgray}\textbf{68.68}} &
  \multicolumn{1}{c|}{\cellcolor{lightgray}\textbf{84.42}} &
  \cellcolor{lightgray}\textbf{67.48} &
  \multicolumn{1}{c|}{\cellcolor{lightgray}\textbf{84.42}} &
  \multicolumn{1}{c|}{\cellcolor{lightgray}\textbf{96.28}} &
  \cellcolor{lightgray}\textbf{74.49} \\
 &
  DAAN\cite{DAAN} &
  \multicolumn{1}{c|}{68.26} &
  \multicolumn{1}{c|}{84.21} &
  67.48 &
  \multicolumn{1}{c|}{79.06} &
  \multicolumn{1}{c|}{94.57} &
  65.63 \\
 &
  \cellcolor{lightgray}\textbf{+MCRL(Ours)} &
  \multicolumn{1}{c|}{\cellcolor{lightgray}\textbf{68.84}} &
  \multicolumn{1}{c|}{\cellcolor{lightgray}\textbf{84.83}} &
  \cellcolor{lightgray}\textbf{68.03} &
  \multicolumn{1}{c|}{\cellcolor{lightgray}\textbf{80.65}} &
  \multicolumn{1}{c|}{\cellcolor{lightgray}\textbf{95.46}} &
  \cellcolor{lightgray}\textbf{67.06} \\ \hline
\multirow{4}{*}{ResNet101} &
  DSAN\cite{DSAN} &
  \multicolumn{1}{c|}{68.04} &
  \multicolumn{1}{c|}{84.51} &
  66.38 &
  \multicolumn{1}{c|}{81.53} &
  \multicolumn{1}{c|}{95.99} &
  72.55 \\
 &
  \cellcolor{lightgray}\textbf{+MCRL(Ours)} &
  \multicolumn{1}{c|}{\cellcolor{lightgray}\textbf{69.31}} &
  \multicolumn{1}{c|}{\cellcolor{lightgray}\textbf{84.84}} &
  \cellcolor{lightgray}\textbf{68.36} &
  \multicolumn{1}{c|}{\cellcolor{lightgray}81.47} &
  \multicolumn{1}{c|}{\cellcolor{lightgray}95.69} &
  \cellcolor{lightgray}72.21 \\
 &
  DAAN\cite{DAAN} &
  \multicolumn{1}{c|}{68.48} &
  \multicolumn{1}{c|}{84.54} &
  67.51 &
  \multicolumn{1}{c|}{80.00} &
  \multicolumn{1}{c|}{95.75} &
  72.05 \\
 &
  \cellcolor{lightgray}\textbf{+MCRL(Ours)} &
  \multicolumn{1}{c|}{\cellcolor{lightgray}\textbf{68.66}} &
  \multicolumn{1}{c|}{\cellcolor{lightgray}\textbf{84.79}} &
  \cellcolor{lightgray}\textbf{67.82} &
  \multicolumn{1}{c|}{\cellcolor{lightgray}81.88} &
  \multicolumn{1}{c|}{\cellcolor{lightgray}96.76} &
  \cellcolor{lightgray}73.64 \\ \hline
\multirow{4}{*}{ResNet152} &
  DSAN\cite{DSAN} &
  \multicolumn{1}{c|}{68.25} &
  \multicolumn{1}{c|}{84.48} &
  66.34 &
  \multicolumn{1}{c|}{80.94} &
  \multicolumn{1}{c|}{94.40} &
  72.35 \\
 &
  \cellcolor{lightgray}\textbf{+MCRL(Ours)} &
  \multicolumn{1}{c|}{\cellcolor{lightgray}\textbf{69.64}} &
  \multicolumn{1}{c|}{\cellcolor{lightgray}\textbf{85.47}} &
  \cellcolor{lightgray}\textbf{68.69} &
  \multicolumn{1}{c|}{\cellcolor{lightgray}\textbf{82.30}} &
  \multicolumn{1}{c|}{\cellcolor{lightgray}\textbf{95.51}} &
  \cellcolor{lightgray}\textbf{72.72} \\
 &
  DAAN\cite{DAAN} &
  \multicolumn{1}{c|}{69.22} &
  \multicolumn{1}{c|}{84.88} &
  68.40 &
  \multicolumn{1}{c|}{81.24} &
  \multicolumn{1}{c|}{94.93} &
  70.20 \\
 &
  \cellcolor{lightgray}\textbf{+MCRL(Ours)} &
  \multicolumn{1}{c|}{\cellcolor{lightgray}\textbf{69.44}} &
  \multicolumn{1}{c|}{\cellcolor{lightgray}\textbf{85.01}} &
  \cellcolor{lightgray}\textbf{68.55} &
  \multicolumn{1}{c|}{\cellcolor{lightgray}81.17} &
  \multicolumn{1}{c|}{\cellcolor{lightgray}\textbf{95.10}} &
  \cellcolor{lightgray}\textbf{71.18} \\ \hline
\multirow{4}{*}{ViT} &
  CDTrans(DeiT-S)\cite{CDTrans} &
  \multicolumn{1}{c|}{79.00} &
  \multicolumn{1}{c|}{90.52} &
  78.31 &
  \multicolumn{1}{c|}{90.09} &
  \multicolumn{1}{c|}{96.34} &
  82.97 \\
 &
  \cellcolor{lightgray}\textbf{+MCRL(Ours)} &
  \multicolumn{1}{c|}{\cellcolor{lightgray}79.00} &
  \multicolumn{1}{c|}{\cellcolor{lightgray}\textbf{91.00}} &
  \cellcolor{lightgray}77.95 &
  \multicolumn{1}{c|}{\cellcolor{lightgray}\textbf{93.20}} &
  \multicolumn{1}{c|}{\cellcolor{lightgray}\textbf{97.94}} &
  \cellcolor{lightgray}\textbf{85.38} \\ 
 &
  CDTrans(DeiT-B)\cite{CDTrans} &
  \multicolumn{1}{c|}{81.93} &
  \multicolumn{1}{c|}{92.55} &
  81.43 &
  \multicolumn{1}{c|}{90.86} &
  \multicolumn{1}{c|}{97.35} &
  83.35 \\
 &
  \cellcolor{lightgray}\textbf{+MCRL(Ours)} &
  \multicolumn{1}{c|}{\cellcolor{lightgray}\textbf{82.10}} &
  \multicolumn{1}{c|}{\cellcolor{lightgray}\textbf{93.03}} &
  \cellcolor{lightgray}\textbf{81.46} &
  \multicolumn{1}{c|}{\cellcolor{lightgray}\textbf{93.63}} &
  \multicolumn{1}{c|}{\cellcolor{lightgray}\textbf{98.47}} &
  \cellcolor{lightgray}\textbf{88.38} \\ \hline
\end{tabular}
\vspace{-0.1in}
\end{table*}

\subsection{Ablation Study}
In this section, we present a comprehensive ablation study to evaluate the contribution of MCRL and investigate its influence on overall performance. By systematically analyzing and comparing the results, we aim to gain deeper insights into the key factors that contribute to the success of our proposed approach. We compared three distinct implementations of multi-label methods on the DailyFood-16 dataset. All these methods can be seen as special cases of our most efficacious Multi-Cluster Reference Learning via soft-label strategy. 

\subsubsection{Ratio Adjustment Method(RAM)}
We extract the logits of the two highest-confidence categories for each target sample after $f_\theta$ and calculate their $ratio$. If the $ratio$ exceeds a predefined threshold, it signifies that the feature of the target sample is closer to the feature of the source domain from the same category with the highest confidence. In such scenarios, we only use the feature of the highest confidence category of the source domain, minimizing the distance between it and the feature distribution of the target sample. Conversely, if the $ratio$ is below the specified threshold, it implies that the target sample's feature closely resembles the features of the two categories with the highest confidence. In such cases, we utilize the features of the top two categories, reducing the distance between their feature distributions and the feature distribution of the target sample. This design rationale acknowledges the fact that a single image of food may contain multiple potential primary ingredients, which could originate from 2 dishes. Consequently, we assign up to 2 possible categories to each food image to adaptively align with the source domain.

\subsubsection{Hard-Selection Method(HM)}
Considering that a single food image may closely align with the features of more than two categories of food, for instance, a dish containing cucumber, tofu skin, and black fungus, it is challenging to determine whether it belongs to ``Marinated cold cucumber", ``Yuba salad", or ``Black fungus salad". In such cases, we aim to draw the feature distribution of this image closer to the feature distributions of all three categories in the source domain. Therefore, we set different values for $n$, where $n$ represents how many classes of source domain features each target domain sample needs to align in the feature space. In this case, we give the same weight to $n$ classes.

\subsubsection{Soft-Selection Method(SM)}
We adopt SM, which uses MCRL to align the source and target domains, wherein each class of samples is assigned varying weights based on classification confidence to approximate the source domain.

The results on each dataset in TABLE \ref{Ablation} indicate that it is not enough to align them by using PAM, or aligning them with equal weights. Therefore, the proposed MCRL is able to perform dynamic distribution alignment between domains and achieve better performance on food recognition. This property of MCRL is extremely important in real applications since given an unknown target domain, we can never know the specific number of classes in the source domain whose features more closely resemble the target sample. MCRL makes it possible to easily, dynamically, and quantitatively evaluate their relative importance in everyday food recognition.
\vspace{-0.1in}

\begin{table}[]
\centering
\caption{Ablation study of multi-label on DailyFood-16.}
\label{Ablation}
\begin{tabular}{cc|c|c}
\hline
\multicolumn{2}{c|}{Methods}    & DSAN\cite{DSAN} & CDTrans\cite{CDTrans} \\ \hline
\multicolumn{2}{c|}{Baselines}         & 82.24 & 90.86 \\ \hline
\multicolumn{1}{c|}{}    & $ratio=1.1$ & 83.01 & 93.22 \\
\multicolumn{1}{c|}{RAM} & $ratio=1.2$ & 82.48 & 92.92 \\
\multicolumn{1}{c|}{}    & $ratio=1.5$ & 82.83 & 93.2  \\ \hline
\multicolumn{1}{c|}{}    & $k=2$       & 83.95 & 93.3  \\
\multicolumn{1}{c|}{HM}  & $k=3$       & 83.89 & 93.6  \\
\multicolumn{1}{c|}{}    & $k=4$       & 82.77 & 90.9  \\ \hline
\multicolumn{1}{c|}{}    & $k=2$       & 83.07 & 93.27 \\
\multicolumn{1}{c|}{SM} & $k=3$ & \textbf{84.42}  & \textbf{93.63}        \\
\multicolumn{1}{c|}{}    & $k=4$       & 83.01 & 93.51 \\ \hline
\end{tabular}
\vspace{-0.1in}
\end{table}

\subsection{Qualitative Results}
\begin{figure}[!t]
\centering
\includegraphics[width=3.2in]{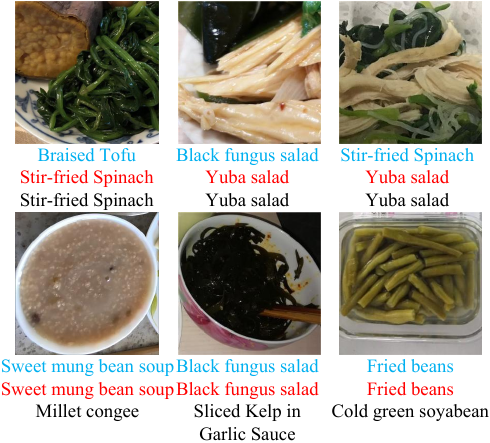}
\caption{Qualitative examples on DailyFood-16 dataset. For each image, the three lines of captions below indicate the predictions of CDTrans (blue), the predictions of MCRL (red), and the ground truth (black), respectively.}
\label{case}
\vspace{-0.15in}
\end{figure}

Food classification is a complex task, particularly for data wherein the distribution of source and target domain samples diverges, as it relies heavily on the model's capability to extract the correct features corresponding to food categories. Fig. \ref{case} presents examples of correct and incorrect classifications by our model when the target domain is the DailyFood-16 dataset. For the successful samples, we show SOTA method CDTrans \cite{CDTrans} fail but our MCRL could make it. As shown in Fig. \ref{case}, images in the top row are misclassified by the SOTA method CDTrans but correctly classified by our method MCRL. In the second row of the Fig. \ref{case}, we present the samples where MCRL made incorrect predictions. These images often pose challenges in accurate classification due to issues such as dim lighting, cluttered ingredients, and subtle appearance differences, making them difficult to be clearly distinguished even by human vision.
In summary, our model may misclassify images where the inherent features are not clear, or images that contain distinctive characteristics of other food categories. However, these errors usually result from the nature of the images themselves, as even human vision would find it challenging to accurately distinguish these pictures.
\vspace{-0.1in}


\subsection{Discussion}
 While previous UDA methods adopt an end-to-end training strategy that simultaneously train the classifier and perform domain adaptation, we use a two-stage process here. Specifically, we first train the classifier and subsequently conduct domain adaptation, which we refer to as the "2-stage" approach. The results are shown in Tab. \ref{2stage}. It indicates that the pseudo-labels assigned to target domain samples, acquired post-training the model on the source domain, are more accurate. 
 Hence, category-level Unsupervised Domain Adaptation (UDA) methods can extract more robust features, enhancing the classification of food images. 

In addition, considering the similarity among images from the same class in our three datasets: VIREO Food-172, DailyFood-172, and DailyFood-16, we leverage an alternate target domain as an intermediate domain for training on DailyFood-172 and DailyFood-16 datasets to enlarge the dataset in order to enhance the generalization capability of the model. Specifically, VIREO Food-172 consistently serves as the source domain. Initially, we train using DailyFood-16 as the target domain and save the model weights, then switch the target domain to DailyFood-172, and re-train using the pre-trained model to obtain the final results. The experiment results are presented in the ``intermediate domain" section of TABLE \ref{2stage}. The results indicate that due to the similarities between images of the same categories of food, the neural network was able to learn more features, demonstrating the effective mutual assistance between the two target domains. It's worth noting that by employing the MCRL method, we have surpassed the baseline in all three settings, thus positioning ourselves as the leading-edge approach. To detail further, ResNet50 was employed as the backbone in all the aforementioned experiments and Deit-S is used in CDTrans.

\begin{table}[]
\centering
\caption{Experimental results of the 2-stage and Enlarged Dataset settings.}
\label{2stage}
\begin{tabular}{c|c}
\hline
Methods                        & DailyFood-172(\%) \\ \hline
DSAN\cite{DSAN}                           & 66.96             \\
DSAN-2-stage\cite{DSAN}                   & 68.42             \\
DSAN-Enlarged Dataset\cite{DSAN}       & 67.26             \\
DSAN+MCRL\cite{DSAN}                     & 68.80             \\
DSAN-2-stage+MCRL\cite{DSAN}             & \textbf{69.51}    \\
DSAN-Enlarged Dataset+MCRL\cite{DSAN} & 68.57             \\ \hline
DAAN\cite{DAAN}                           & 68.26             \\
DAAN-2-stage\cite{DAAN}                   & 68.78             \\
DAAN-Enlarged Dataset\cite{DAAN}       & 68.29             \\
DAAN+MCRL\cite{DAAN}                     & 68.84             \\
DAAN-2-stage+MCRL\cite{DAAN}             & \textbf{69.31}    \\
DAAN-Enlarged Dataset+MCRL\cite{DAAN} & 68.70             \\ \hline
CDTrans\cite{CDTrans}                        & 78.98             \\
CDTrans-Enlarged Dataset\cite{CDTrans}    & 78.94             \\
CDTrans+MCRL\cite{CDTrans}                  & \textbf{79.00}    \\ \hline
\end{tabular}
\end{table}


\section{CONCLUSIONS}
In this paper, we presented two novel food datasets DailyFood-172 and DailyFood-16, along with a proposed baseline method called Multi-Cluster Reference Learning (MCRL). The DailyFood-172 dataset is collected from regular individuals who uploaded their home-cooked dishes to a website following recipes, whereas the DailyFood-16 dataset comes from daily meals uploaded to a website by people on a weight loss regime, acquired through searches and cropping of the images. And the proposed MCRL learns the domain shift between a certain target sample and multiple source clusters. Experimental results on extensive visual cross-domain tasks have shown that applying our method to conventional UDA methods can improve the classification accuracy of the target domain and further lead to more discriminative and domain invariant features than the conventional UDA baselines.


\bibliographystyle{IEEEtran}
\bibliography{IEEEfull}

\begin{thebibliography}{10}
\providecommand{\url}[1]{#1}
\csname url@samestyle\endcsname
\providecommand{\newblock}{\relax}
\providecommand{\bibinfo}[2]{#2}
\providecommand{\BIBentrySTDinterwordspacing}{\spaceskip=0pt\relax}
\providecommand{\BIBentryALTinterwordstretchfactor}{4}
\providecommand{\BIBentryALTinterwordspacing}{\spaceskip=\fontdimen2\font plus
\BIBentryALTinterwordstretchfactor\fontdimen3\font minus
  \fontdimen4\font\relax}
\providecommand{\BIBforeignlanguage}[2]{{%
\expandafter\ifx\csname l@#1\endcsname\relax
\typeout{** WARNING: IEEEtran.bst: No hyphenation pattern has been}%
\typeout{** loaded for the language `#1'. Using the pattern for}%
\typeout{** the default language instead.}%
\else
\language=\csname l@#1\endcsname
\fi
#2}}
\providecommand{\BIBdecl}{\relax}
\BIBdecl

\bibitem{oxford}
W.~Min, Z.~Wang, Y.~Liu, M.~Luo, L.~Kang, X.~Wei, X.~Wei, and S.~Jiang, ``Large
  scale visual food recognition,'' \emph{CoRR}, vol. abs/2103.16107, 2021.

\bibitem{Cedric}
Q.~Thames, A.~Karpur, W.~Norris, F.~Xia, L.~Panait, T.~Weyand, and J.~Sim,
  ``Nutrition5k: Towards automatic nutritional understanding of generic food,''
  in \emph{IEEE Conference on Computer Vision and Pattern Recognition}, 2021,
  pp. 8903--8911.

\bibitem{Image}
Y.~Liang, J.~Li, Q.~Zhao, W.~Rao, C.~Zhang, and C.~Wang, ``Image segmentation
  and recognition for multi-class chinese food,'' in \emph{International
  Conference on Image Processing}, 2022, pp. 3938--3942.

\bibitem{Deep-based}
J.~Chen and C.-W. Ngo, ``Deep-based ingredient recognition for cooking recipe
  retrieval,'' in \emph{ACM Multimedia}, 2016, pp. 32--41.

\bibitem{food101}
L.~Bossard, M.~Guillaumin, and L.~Van~Gool, ``Food-101--mining discriminative
  components with random forests,'' in \emph{European Conference on Computer
  Vision}.\hskip 1em plus 0.5em minus 0.4em\relax Springer, 2014, pp. 446--461.

\bibitem{Region-Wise}
J.~Chen, B.~Zhu, C.-W. Ngo, T.-S. Chua, and Y.-G. Jiang, ``A study of
  multi-task and region-wise deep learning for food ingredient recognition,''
  \emph{IEEE Trans. on Image Processing}, vol.~30, pp. 1514--1526, 2020.

\bibitem{MMD}
E.~Tzeng, J.~Hoffman, N.~Zhang, K.~Saenko, and T.~Darrell, ``Deep domain
  confusion: Maximizing for domain invariance,'' \emph{arXiv preprint
  arXiv:1412.3474}, 2014.

\bibitem{DAN}
M.~Long, Y.~Cao, J.~Wang, and M.~Jordan, ``Learning transferable features with
  deep adaptation networks,'' in \emph{International Conference on Machine
  Learning}.\hskip 1em plus 0.5em minus 0.4em\relax PMLR, 2015, pp. 97--105.

\bibitem{CORAL}
B.~Sun and K.~Saenko, ``Deep coral: Correlation alignment for deep domain
  adaptation,'' in \emph{ECCV Workshop}.\hskip 1em plus 0.5em minus 0.4em\relax
  Springer, 2016, pp. 443--450.

\bibitem{DSN}
K.~Bousmalis, G.~Trigeorgis, N.~Silberman, D.~Krishnan, and D.~Erhan, ``Domain
  separation networks,'' \emph{Advances in neural information processing
  systems}, vol.~29, 2016.

\bibitem{Similarity-based}
M.~Peng, Z.~Li, and X.~Juan, ``Similarity-based domain adaptation network,''
  \emph{Neurocomputing}, vol. 493, pp. 462--473, 2022.

\bibitem{CCUDA}
J.~Huang, D.~Guan, A.~Xiao, S.~Lu, and L.~Shao, ``Category contrast for
  unsupervised domain adaptation in visual tasks,'' in \emph{IEEE Conference on
  Computer Vision and Pattern Recognition}, 2022, pp. 1203--1214.

\bibitem{DSAN}
Y.~Zhu, F.~Zhuang, J.~Wang, G.~Ke, J.~Chen, J.~Bian, H.~Xiong, and Q.~He,
  ``Deep subdomain adaptation network for image classification,'' \emph{IEEE
  Trans. on Neural Networks and Learning Systems}, vol.~32, no.~4, pp.
  1713--1722, 2020.

\bibitem{DAAN}
C.~Yu, J.~Wang, Y.~Chen, and M.~Huang, ``Transfer learning with dynamic
  adversarial adaptation network,'' in \emph{IEEE International Conference on
  Data Mining (ICDM)}.\hskip 1em plus 0.5em minus 0.4em\relax IEEE, 2019, pp.
  778--786.

\bibitem{CAN}
G.~Kang, L.~Jiang, Y.~Yang, and A.~G. Hauptmann, ``Contrastive adaptation
  network for unsupervised domain adaptation,'' in \emph{IEEE Conference on
  Computer Vision and Pattern Recognition}, 2019, pp. 4893--4902.

\bibitem{pseudo}
J.~Wang and X.-L. Zhang, ``Improving pseudo labels with intra-class similarity
  for unsupervised domain adaptation,'' \emph{Pattern Recognition}, vol. 138,
  p. 109379, 2023.

\bibitem{DADA}
Y.~Du, Z.~Tan, Q.~Chen, X.~Zhang, Y.~Yao, and C.~Wang, ``Dual adversarial
  domain adaptation,'' \emph{arXiv preprint arXiv:2001.00153}, 2020.

\bibitem{MEDA}
J.~Wang, W.~Feng, Y.~Chen, H.~Yu, M.~Huang, and P.~S. Yu, ``Visual domain
  adaptation with manifold embedded distribution alignment,'' in \emph{ACM
  Multimedia}, 2018, pp. 402--410.

\bibitem{JDA}
M.~Long, J.~Wang, G.~Ding, J.~Sun, and P.~S. Yu, ``Transfer feature learning
  with joint distribution adaptation,'' in \emph{IEEE International Conference
  on Computer Vision}, 2013, pp. 2200--2207.

\bibitem{FoodX-251}
P.~Kaur, K.~Sikka, W.~Wang, S.~Belongie, and A.~Divakaran, ``Foodx-251: a
  dataset for fine-grained food classification,'' \emph{arXiv preprint
  arXiv:1907.06167}, 2019.

\bibitem{Cooking}
A.~Salvador, N.~Hynes, Y.~Aytar, J.~Marin, F.~Ofli, I.~Weber, and A.~Torralba,
  ``Learning cross-modal embeddings for cooking recipes and food images,'' in
  \emph{IEEE Conference on Computer Vision and Pattern Recognition}, 2017, pp.
  3020--3028.

\bibitem{Automatic}
M.-Y. Chen, Y.-H. Yang, C.-J. Ho, S.-H. Wang, S.-M. Liu, E.~Chang, C.-H. Yeh,
  and M.~Ouhyoung, ``Automatic chinese food identification and quantity
  estimation,'' in \emph{SIGGRAPH Asia 2012 Technical Briefs}, 2012, pp. 1--4.

\bibitem{Recognition}
M.~Puri, Z.~Zhu, Q.~Yu, A.~Divakaran, and H.~Sawhney, ``Recognition and volume
  estimation of food intake using a mobile device,'' in \emph{Workshop on
  Applications of Computer Vision}, 2009, pp. 1--8.

\bibitem{Multiple-food}
Y.~Matsuda and K.~Yanai, ``Multiple-food recognition considering co-occurrence
  employing manifold ranking,'' in \emph{International Conference on Pattern
  Recognition}.\hskip 1em plus 0.5em minus 0.4em\relax IEEE, 2012, pp.
  2017--2020.

\bibitem{FOODCLASSIFICATION}
A.~{\c{S}}eng{\"u}r, Y.~Akbulut, and {\"U}.~Budak, ``Food image classification
  with deep features,'' in \emph{International Artificial Intelligence and Data
  Processing Symposium (IDAP)}.\hskip 1em plus 0.5em minus 0.4em\relax Ieee,
  2019, pp. 1--6.

\bibitem{Database}
B.~Arslan, S.~Memi{\c{s}}, E.~B. S{\"o}nmez, and O.~Z. Batur, ``Fine-grained
  food classification methods on the uec food-100 database,'' \emph{IEEE
  Transactions on Artificial Intelligence}, vol.~3, no.~2, pp. 238--243, 2021.

\bibitem{food-transfer}
B.~Zhu, C.-W. Ngo, and J.-j. Chen, ``Cross-domain cross-modal food transfer,''
  in \emph{ACM Multimedia}, 2020, pp. 3762--3770.

\bibitem{DANN}
Y.~Ganin, E.~Ustinova, H.~Ajakan, P.~Germain, H.~Larochelle, F.~Laviolette,
  M.~Marchand, and V.~Lempitsky, ``Domain-adversarial training of neural
  networks,'' \emph{The journal of machine learning research}, vol.~17, no.~1,
  pp. 2096--2030, 2016.

\bibitem{Return}
B.~Sun, J.~Feng, and K.~Saenko, ``Return of frustratingly easy domain
  adaptation,'' in \emph{AAAI conference on artificial intelligence}, vol.~30,
  no.~1, 2016.

\bibitem{Ghifary}
M.~Ghifary, W.~B. Kleijn, and M.~Zhang, ``Domain adaptive neural networks for
  object recognition,'' in \emph{Pacific Rim International Conference on
  Artificial Intelligence}.\hskip 1em plus 0.5em minus 0.4em\relax Springer,
  2014, pp. 898--904.

\bibitem{JAN}
M.~Long, H.~Zhu, J.~Wang, and M.~I. Jordan, ``Deep transfer learning with joint
  adaptation networks,'' in \emph{International Conference on Machine
  Learning}.\hskip 1em plus 0.5em minus 0.4em\relax PMLR, 2017, pp. 2208--2217.

\bibitem{GAN}
I.~Goodfellow, J.~Pouget-Abadie, M.~Mirza, B.~Xu, D.~Warde-Farley, S.~Ozair,
  A.~Courville, and Y.~Bengio, ``Generative adversarial networks,''
  \emph{Communications of the ACM}, vol.~63, no.~11, pp. 139--144, 2020.

\bibitem{Backpropagation}
Y.~Ganin and V.~Lempitsky, ``Unsupervised domain adaptation by
  backpropagation,'' in \emph{International Conference on Machine
  Learning}.\hskip 1em plus 0.5em minus 0.4em\relax PMLR, 2015, pp. 1180--1189.

\bibitem{Dlow}
R.~Gong, W.~Li, Y.~Chen, and L.~V. Gool, ``Dlow: Domain flow for adaptation and
  generalization,'' in \emph{IEEE Conference on Computer Vision and Pattern
  Recognition}, 2019, pp. 2477--2486.

\bibitem{Cycada}
J.~Hoffman, E.~Tzeng, T.~Park, J.-Y. Zhu, P.~Isola, K.~Saenko, A.~Efros, and
  T.~Darrell, ``Cycada: Cycle-consistent adversarial domain adaptation,'' in
  \emph{International Conference on Machine Learning}.\hskip 1em plus 0.5em
  minus 0.4em\relax Pmlr, 2018, pp. 1989--1998.

\bibitem{DTN}
X.~Zhang, F.~X. Yu, S.-F. Chang, and S.~Wang, ``Deep transfer network:
  Unsupervised domain adaptation,'' \emph{arXiv preprint arXiv:1503.00591},
  2015.

\bibitem{MADA}
Z.~Pei, Z.~Cao, M.~Long, and J.~Wang, ``Multi-adversarial domain adaptation,''
  in \emph{AAAI conference on artificial intelligence}, vol.~32, no.~1, 2018.

\bibitem{Pseudo-label}
D.-H. Lee \emph{et~al.}, ``Pseudo-label: The simple and efficient
  semi-supervised learning method for deep neural networks,'' in \emph{Workshop
  on challenges in representation learning, ICML}, vol.~3, no.~2, 2013, p. 896.

\bibitem{Spherical}
X.~Gu, J.~Sun, and Z.~Xu, ``Spherical space domain adaptation with robust
  pseudo-label loss,'' in \emph{IEEE Conference on Computer Vision and Pattern
  Recognition}, 2020, pp. 9101--9110.

\bibitem{Fisher}
Y.~Zhang, Y.~Zhang, Y.~Wei, K.~Bai, Y.~Song, and Q.~Yang, ``Fisher deep domain
  adaptation,'' in \emph{International Conference on Data Mining}.\hskip 1em
  plus 0.5em minus 0.4em\relax SIAM, 2020, pp. 469--477.

\bibitem{Rectifying}
Z.~Zheng and Y.~Yang, ``Rectifying pseudo label learning via uncertainty
  estimation for domain adaptive semantic segmentation,'' \emph{International
  Journal of Computer Vision}, vol. 129, no.~4, pp. 1106--1120, 2021.

\bibitem{kernel}
A.~Gretton, K.~M. Borgwardt, M.~J. Rasch, B.~Sch{\"o}lkopf, and A.~Smola, ``A
  kernel two-sample test,'' \emph{The Journal of Machine Learning Research},
  vol.~13, no.~1, pp. 723--773, 2012.

\bibitem{qin2019rethinking}
Z.~Qin, D.~Kim, and T.~Gedeon, ``Rethinking softmax with cross-entropy: Neural
  network classifier as mutual information estimator,'' \emph{arXiv preprint
  arXiv:1911.10688}, 2019.

\bibitem{officehome}
H.~Venkateswara, J.~Eusebio, S.~Chakraborty, and S.~Panchanathan, ``Deep
  hashing network for unsupervised domain adaptation,'' in \emph{IEEE
  Conference on Computer Vision and Pattern Recognition}, 2017, pp. 5018--5027.

\bibitem{VISDA}
X.~Peng, B.~Usman, N.~Kaushik, J.~Hoffman, D.~Wang, and K.~Saenko, ``Visda: The
  visual domain adaptation challenge,'' \emph{arXiv preprint arXiv:1710.06924},
  2017.

\bibitem{covi}
J.~Na, D.~Han, H.~J. Chang, and W.~Hwang, ``Contrastive vicinal space for
  unsupervised domain adaptation,'' in \emph{European Conference on Computer
  Vision}.\hskip 1em plus 0.5em minus 0.4em\relax Springer, 2022, pp. 92--110.

\bibitem{CDTrans}
T.~Xu, W.~Chen, P.~Wang, F.~Wang, H.~Li, and R.~Jin, ``Cdtrans: Cross-domain
  transformer for unsupervised domain adaptation,'' \emph{arXiv preprint
  arXiv:2109.06165}, 2021.

\bibitem{imagenet}
O.~Russakovsky, J.~Deng, H.~Su, J.~Krause, S.~Satheesh, S.~Ma, Z.~Huang,
  A.~Karpathy, A.~Khosla, M.~Bernstein \emph{et~al.}, ``Imagenet large scale
  visual recognition challenge,'' \emph{International Journal of Computer
  Vision}, vol. 115, pp. 211--252, 2015.

\bibitem{adam}
D.~P. Kingma and J.~Ba, ``Adam: A method for stochastic optimization,''
  \emph{arXiv preprint arXiv:1412.6980}, 2014.

\bibitem{vit}
A.~Dosovitskiy, L.~Beyer, A.~Kolesnikov, D.~Weissenborn, X.~Zhai,
  T.~Unterthiner, M.~Dehghani, M.~Minderer, G.~Heigold, S.~Gelly \emph{et~al.},
  ``An image is worth 16x16 words: Transformers for image recognition at
  scale,'' \emph{arXiv preprint arXiv:2010.11929}, 2020.

\bibitem{deit}
H.~Touvron, M.~Cord, M.~Douze, F.~Massa, A.~Sablayrolles, and H.~J{\'e}gou,
  ``Training data-efficient image transformers \& distillation through
  attention,'' in \emph{International Conference on Machine Learning}.\hskip
  1em plus 0.5em minus 0.4em\relax PMLR, 2021, pp. 10\,347--10\,357.

\end{thebibliography}
\begin{IEEEbiography}[{\includegraphics[width=0.9in,height=1.25in,clip,keepaspectratio]{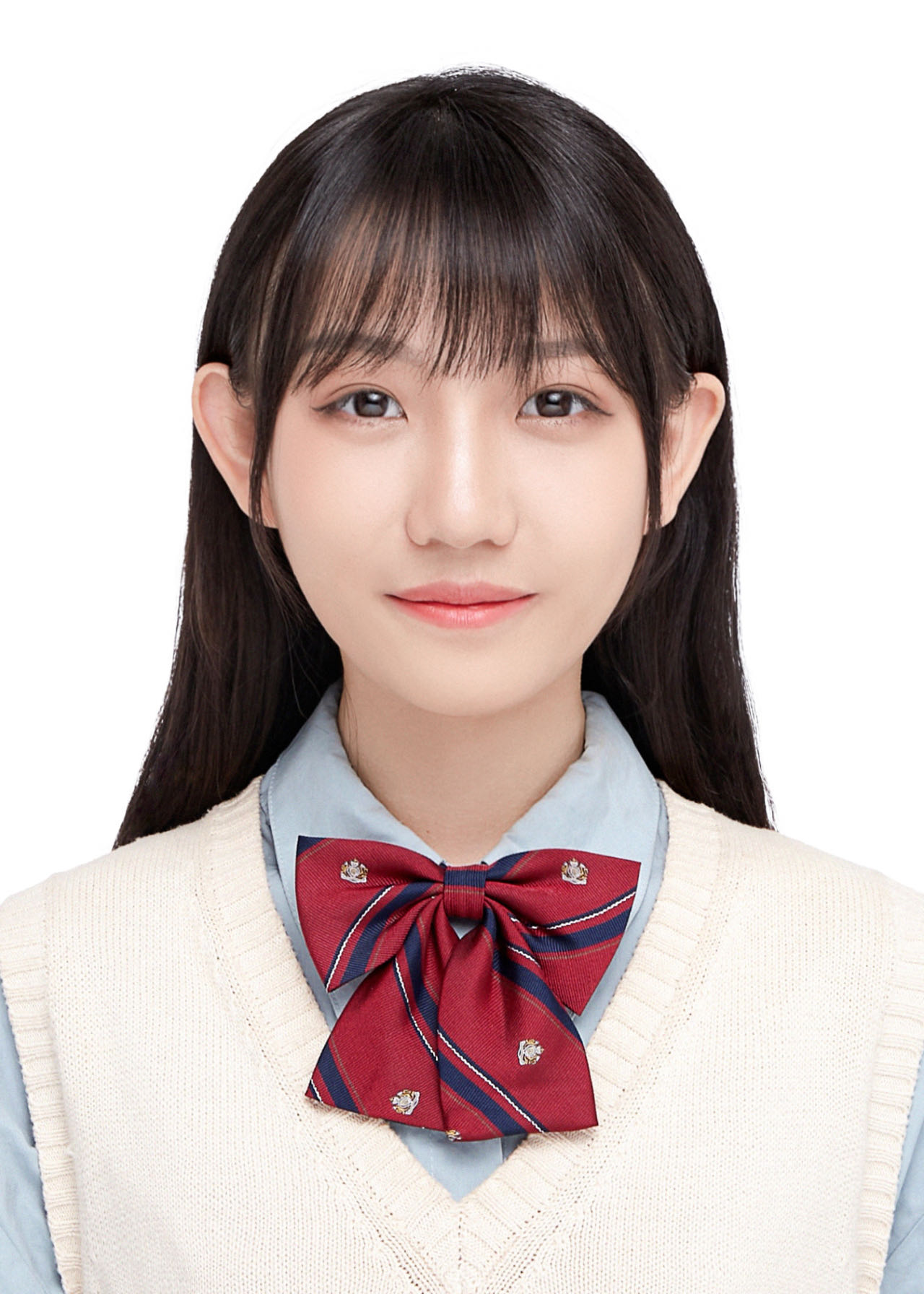}}]{Guoshan Liu} 
received the B.E. degree in Computer Science from Jilin University, Changchun, China, in 2022. She is currently pursuing her M.S. degree in Computer Science at Fudan University. Her research interests include food recognition and domain adaptation.
\end{IEEEbiography}
\begin{IEEEbiography}[{\includegraphics[width=0.9in,height=1.25in,clip,keepaspectratio]{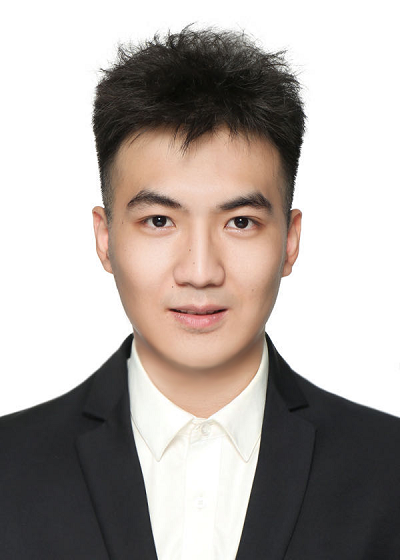}}]{Yang Jiao} 
received the B.E. degree from University of Electronic Science and Technology of China, Chengdu, China, in 2021. He is currently pursuing his Ph.D. degree in Computer Science at Fudan University. His research interests include multi-media analysis and 3D vision.
\end{IEEEbiography}
\begin{IEEEbiography}[{\includegraphics[width=0.9in,height=1.25in,clip,keepaspectratio]{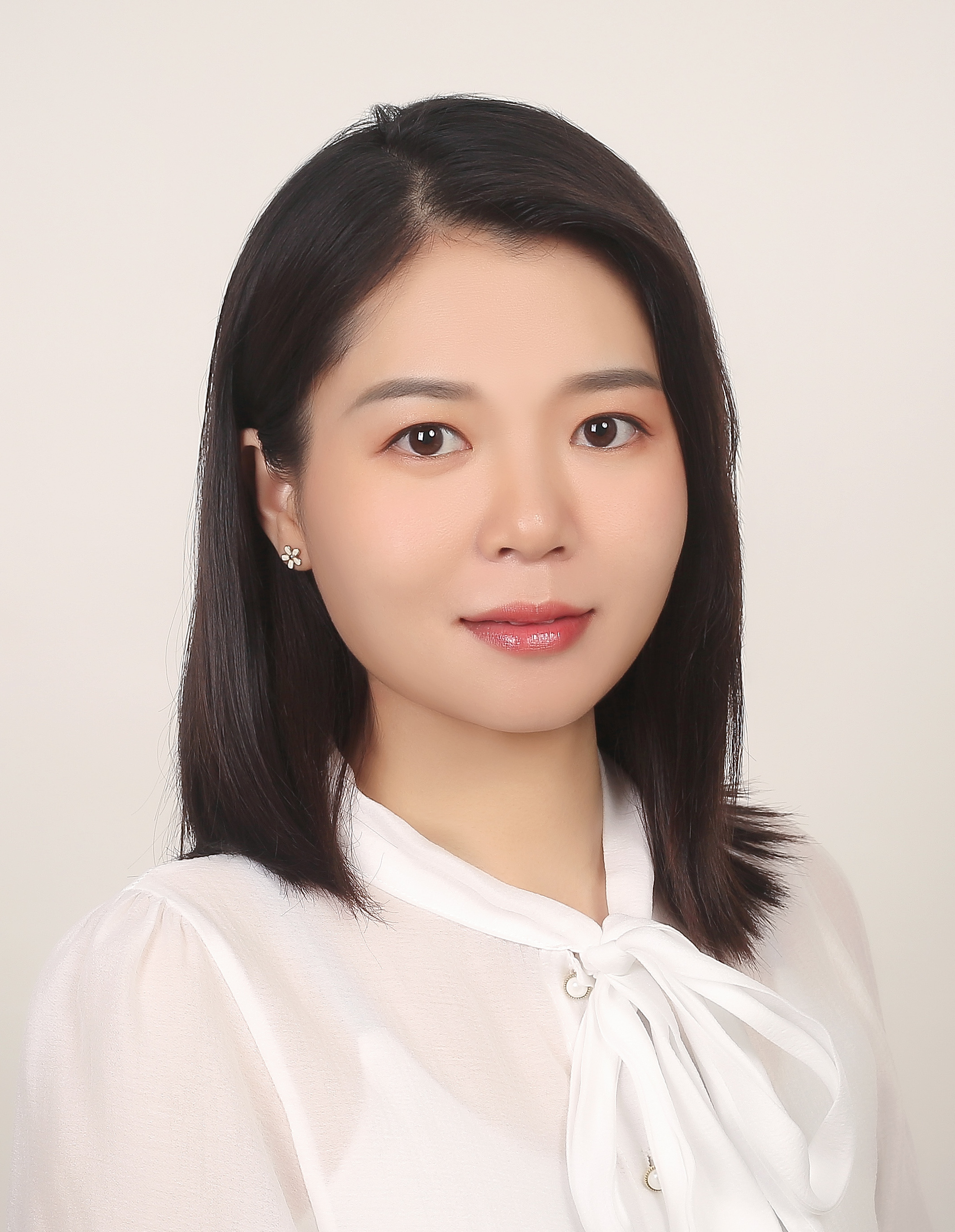}}]{Jingjing Chen} 
is now an Associate Professor at the School of Computer Science, Fudan
University. Before joining Fudan, she
was a postdoc research fellow at the School of
Computing in the National University of Singapore.
She received her Ph.D. degree in Computer Science
from the City University of Hong Kong in 2018.
Her research interest lies in diet tracking and nutrition estimation based on multi-modal processing
of food images, including food recognition, cross-modal recipe retrieval.
\end{IEEEbiography}
\begin{IEEEbiography}[{\includegraphics[width=0.9in,height=1.25in,clip,keepaspectratio]{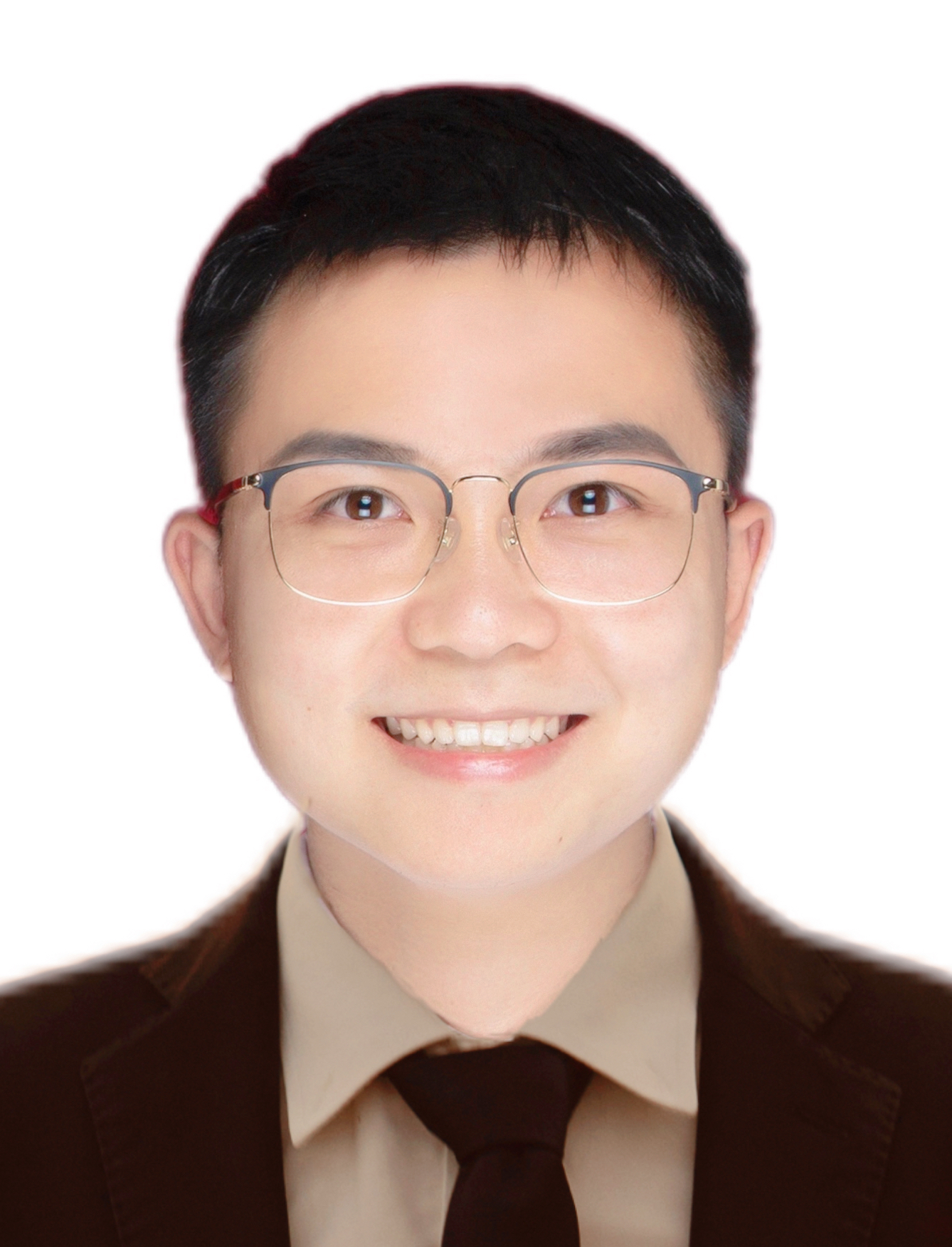}}]{Bin Zhu} received the B.E. degree from Southeast University, Nanjing, China, the M.E. degree from Zhejiang University, Hangzhou, China, and Ph.D. degree from City University of Hong Kong, Hong Kong, China. He is currently an Assistant Professor of Computer Science in the School of Computing and Information Systems , Singapore Management University (SMU). Prior to SMU, He was a postdoctoral researcher in department of computer science, University of Bristol, United Kingdom. His research interests mainly lie in cross-modal recipe retrieval, domain adaptation and egocentric video understanding. Dr. Zhu is the co-organizer of EPIC KITCHENS100 CHALLENGES and co-organizer of Special Session on Multimedia on Cooking and Eating Activities hosted in ACM Multimedia Asia 2023. 
\end{IEEEbiography}
\begin{IEEEbiography}[{\includegraphics[width=0.9in,height=1.25in,clip,keepaspectratio]{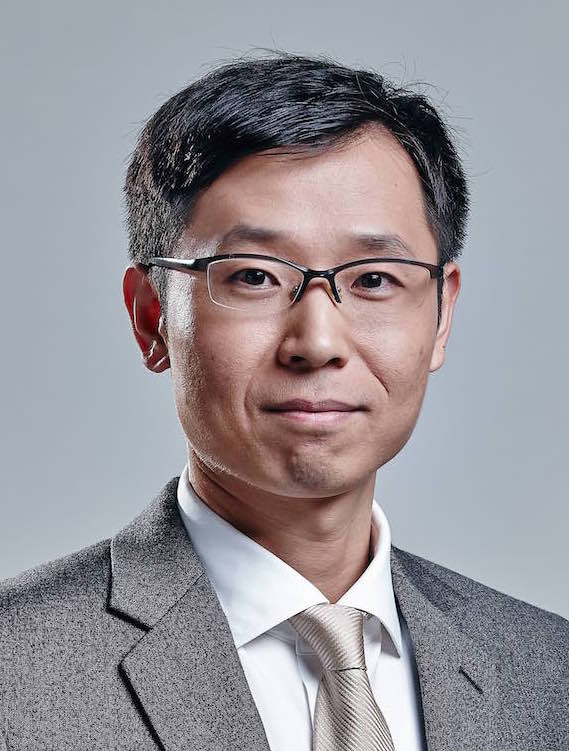}}]{Yu-Gang Jiang} received the Ph.D. degree in Computer Science from City University of Hong Kong in 2009 and worked as a Postdoctoral Research Scientist at Columbia University, New York, during 2009-2011. He is currently a Professor of Computer Science at Fudan University, Shanghai, China. He has been elected as an IEEE Fellow in 2023. His research lies in the areas of multimedia, computer vision, and robust and trustworthy AI. His work has led to many awards, including the inaugural ACM China Rising Star Award, the 2015 ACM SIGMM Rising Star Award, the Research Award for Excellent Young Scholars from NSF China, and the Chang Jiang Distinguished Professorship appointed by Ministry of Education of China.
\end{IEEEbiography}

\end{document}